\documentclass[10pt]{article}

\usepackage[final]{colm2026_conference}
\usepackage{amsmath,amssymb}
\usepackage{booktabs}
\usepackage{graphicx}
\usepackage{array}
\usepackage{tabularx}
\usepackage{url}
\usepackage{xcolor}
\usepackage{tikz}
\usepackage{microtype}
\usepackage{placeins}
\usepackage[hidelinks]{hyperref}
\usetikzlibrary{arrows.meta,positioning}

\newcommand{\ind}[1]{\mathbf{1}[#1]}
\newcommand{\err}{\mathrm{err}}
\newcommand{\lift}{\mathrm{lift}}
\newcommand{\fdp}{\widehat{\mathrm{FDP}}}
\newcommand{\janus}{\textsc{Janus}}

\title{Measurement Under Selection: Decoy-Calibrated Failure \\ Audits for Language Models}

\author{Vyzantinos Repantis \\
The Educational Approaches to \\
Virtual Reality Technologies Lab, \\
University of Ioannina, Greece
\And
Ameya Gawde \\
Meta Platforms
\And
Harshvardhan Singh \\
Meta Platforms
}

\begin{document}
\maketitle
\lhead{Accepted at the AI Measurement Science Workshop at COLM 2026}

\begin{abstract}
Knowing how often a language model fails does not explain where its errors concentrate. When auditors examine many explanations, the strongest observed pattern may arise by chance. We introduce \janus{}, a procedure for checking proposed error patterns before reporting them. \janus{} starts with a fixed list of yes/no properties of the examples being evaluated, such as whether the input is long. For each property, it compares the model's error rates on examples with that property and those without it. To see how large a difference can arise by chance, it repeats this calculation after shuffling the yes/no labels across examples without changing the group sizes. These shuffled properties are called decoys. A pattern is reported only if the size of its error difference meets a threshold set using decoys. On separate held-out examples, the same group must still have the higher error rate and the difference must meet a minimum, which was chosen in advance. In a controlled experiment, where the model must find a code in documents containing tables of staff, projects, and renewal codes, \janus{} confirms five related patterns of higher error rates on tasks requiring more lookups across tables. It also confirms a sixth pattern: lower error rates on examples with the needed information at the ends of the tables. In our samples from the MuSiQue and LongBench v2 public benchmarks, SliceLine finds groups with high error rates, while \janus{} reports no confirmed error patterns for the example properties we chose to test. For comparison, we use standard tests that shuffle errors and account for testing many candidates. With the same holdout check, they confirm two to six controlled patterns, depending on the test and threshold, and none on either benchmark. In simulations with no real error patterns, \janus{} reports false patterns more often than Benjamini--Hochberg, depending on the decoy count; we also tested whether it reports a single pattern deliberately built into simulated data. With our sample sizes and reporting thresholds, \janus{} rarely confirmed the pattern when errors were 5 to 15 percentage points more likely in one group than in the other. Our benchmark audits, therefore, cannot rule out real differences of this size.
\end{abstract}

\section{Introduction}
\label{sec:introduction}

Consider our controlled audit of 160 questions. Each asks the model to find a code in a document containing tables of staff, projects, and renewal codes. The model answers 102 correctly, but this total does not show which kinds of questions are most often answered incorrectly. We examine the number of lookups required, how the question is asked, distracting rows, the position of the needed facts, and table formatting.

We call an interpretable yes/no property of an evaluation example a \emph{descriptor}. For example, \texttt{long\_chain} is true when reaching a code from its project requires three table lookups rather than one. The examples with that property form its \emph{slice}. Across the full audit, the model's error rate is 65\% inside this slice and 7.5\% outside it. Subtracting the second rate from the first gives an error-rate \emph{lift} of 57.5 percentage points.

A large lift is worth investigating, but its size alone does not account for how the pattern was selected. Among many candidate properties, some can show large error differences partly by chance. Selecting the largest observed difference can, therefore, make a pattern appear stronger than it really is. This is \emph{measurement under selection}: the same measurements help us choose a pattern and describe its strength. Slice discovery methods search for groups with high error rates \citep{chung2019slicefinder,sagadeeva2021sliceline,eyuboglu2022domino,deon2022spotlight,wang2023infembed,yu2026mcsd} and LLM-assisted debugging can propose further explanations \citep{ghosh2025ladder,chen2025hibug2}. While such tools provide candidates, our focus is deciding which proposed patterns to report.

\janus{} makes this check explicit in two stages. We divide the audit examples into a \emph{discovery} set for selecting candidates and a separate \emph{holdout} set for checking them. On the discovery set, \janus{} shuffles the descriptors' labels across examples retaining the group sizes. These shuffled descriptors, called \emph{decoys}, provide examples of lifts that can arise by chance. Comparing the sizes of real and decoy lifts sets a threshold for selecting candidates. For each selected candidate, \janus{} recomputes the error-rate difference on the holdout set. A candidate is reported only if both groups have enough examples, the same group still has the higher error rate, and the observed difference meets a minimum chosen in advance.

\janus{} checks descriptors proposed by an auditor or another system. The rule used to choose and define those descriptors must be fixed before examining discovery errors. In the protocol studied here, any errors used to propose descriptors must come from examples outside the discovery and holdout sets.

\paragraph{Contributions.} We formulate the reporting problem, specify the decoy and holdout checks, and evaluate them on one controlled audit and samples from two public benchmarks. We compare with standard permutation tests based on shuffling errors, using the same candidates, data splits, and holdout check. We also evaluate the reporting procedure itself: how often it reports patterns when none are present, and how often it misses a pattern deliberately built into simulated data. These experiments show how \janus{} behaves in the settings we tested. They do not guarantee that the fraction of false findings, averaged across repeated audits, stays below a chosen limit.

\section{Method}
\label{sec:method}

The audit starts with a fixed model, evaluation examples with reference answers, a fixed scoring rule, and a list of candidate descriptors. For each example $i$, let $e_i=1$ if the scoring rule marks the model's answer as incorrect, and $e_i=0$ otherwise. Separately, a descriptor $g$ assigns 1 if a property of the example is present and 0 otherwise.

We divide the examples into discovery and holdout sets and fix the rule for choosing and defining descriptors before examining discovery errors. Any errors used to propose descriptors derive from examples outside the discovery and holdout sets. In our public-benchmark analyses, each numerical cutoff is the median of the corresponding metadata measurement across discovery examples, calculated without using discovery or holdout errors. For example, a question-length descriptor is true when the question has more words than this median. For categorical metadata, we define one descriptor per category seen in discovery. We apply all descriptor definitions unchanged to holdout examples.

\subsection{Measuring an association}

For each descriptor, the numbers of examples inside and outside its slice give its on-slice and off-slice \emph{support}. Its \emph{prevalence} is the fraction of examples inside the slice. To avoid comparisons based on very small groups, we apply the support and prevalence filters specified for each audit in Appendix~\ref{app:implementation}. The remaining $L$ descriptors form the eligible library $\mathcal{G}=\{g_1,\ldots,g_L\}$.

Each group's error rate is the number of answers marked incorrect divided by the number of examples in that group. For descriptor $g$, $\err_{\rm dis}(g=1)$ is this rate inside its slice and $\err_{\rm dis}(g=0)$ is the rate outside it. The subscript $\mathrm{dis}$ indicates that only discovery examples are used. The discovery lift is their difference:
\[
\lift_{\rm dis}(g)=\err_{\rm dis}(g=1)-\err_{\rm dis}(g=0).
\]
A positive lift means a higher observed error rate inside the slice, while a negative lift means a lower observed error rate. We compare the size of the error-rate difference, $|\lift_{\rm dis}(g)|$, regardless of whether the error rate is higher inside or outside the slice, while retaining the sign for the holdout check. A magnitude of $0.10$ means a difference of ten percentage points, not a relative change of ten percent.

\subsection{Constructing chance comparisons}
\label{sec:decoys}

To construct a decoy, we randomly shuffle a descriptor's yes/no values across discovery examples while retaining the model's recorded errors. This preserves both group sizes. We calculate the decoy's lift exactly as for a real descriptor (Figure~\ref{fig:decoy}). We generate $K$ decoys independently, choosing each source descriptor with equal probability from the source library. Same sources may be chosen more than once. In our main audits, the source library is $\mathcal{G}$. We use $K=200$ for the controlled audit and $K=100$ for each public benchmark.

\begin{figure}[t]
\centering
\begin{tikzpicture}[x=0.39cm,y=0.39cm,dot/.style={circle,draw=gray!50,line width=0.35pt,minimum size=0.28cm,inner sep=0pt},onreal/.style={dot,fill=orange!75,draw=orange!80!black},ondecoy/.style={dot,fill=gray!45,draw=gray!60},ttl/.style={font=\small\bfseries,anchor=south}]
\node[ttl] at (4.5,1.15) {Real descriptor};
\foreach \r in {0,...,1} \foreach \c in {0,...,9} \node[dot] at (\c,-\r) {};
\foreach \p in {(1,0),(3,0),(7,0),(0,-1),(5,-1),(8,-1)} \node[onreal] at \p {};
\begin{scope}[shift={(13,0)}]
\node[ttl] at (4.5,1.15) {Decoy descriptor};
\foreach \r in {0,...,1} \foreach \c in {0,...,9} \node[dot] at (\c,-\r) {};
\foreach \p in {(2,0),(6,0),(9,0),(1,-1),(4,-1),(7,-1)} \node[ondecoy] at \p {};
\end{scope}
\end{tikzpicture}
\caption{The same 20 discovery examples appear in the same positions in both panels. Filled circles indicate a yes label. Shuffling preserves the six yes labels while keeping each example's recorded error unchanged.}
\label{fig:decoy}
\end{figure}

Shuffling descriptors separately preserves each source's prevalence, but not which properties occur together. Section~\ref{sec:permutation-baselines} also examines shuffling errors across examples while keeping all descriptor labels fixed, preserving those relationships. Interpreting either shuffle as a chance comparison requires assumptions about how the examples were sampled.

\subsection{Procedure A: choosing one discovery threshold}
\label{sec:procedure-a}

At a proposed threshold $\tau$, we ask how many real descriptors would pass and how many decoys would pass the same rule. Let $\mathcal{D}$ be the set of $K$ decoys. The two counts are
\[
R(\tau)=\#\{g\in\mathcal{G}:|\lift_{\rm dis}(g)|\geq\tau\},\qquad D(\tau)=\#\{d\in\mathcal{D}:|\lift_{\rm dis}(d)|\geq\tau\}.
\]
The real and decoy libraries can differ in size. $L$ is the number of eligible real descriptors and $K$ is the number of decoys. The fraction of decoys passing the threshold is $D(\tau)/K$. Multiplying this fraction by $L$ estimates how many selections among real descriptors could arise by chance. We divide this estimate by the number of real descriptors that pass:
\[
\fdp(\tau)=\frac{(L/K)D(\tau)}{\max(1,R(\tau))}.
\]
We call this the estimated \emph{false discovery proportion} (FDP), not a guaranteed bound on false findings. The $\max(1,R(\tau))$ term prevents division by zero. We consider the size of each real descriptor's discovery lift as a possible threshold. These values form the set of candidate thresholds, $\mathcal{T}=\{|\lift_{\rm dis}(g)|:g\in\mathcal{G}\}$. We choose a limit $q$ on the estimated false discovery proportion. Then, Procedure A selects the smallest lift threshold that meets this limit, retaining the most descriptors allowed by the rule:
\[
\tau^*=\min\{\tau\in\mathcal{T}:\fdp(\tau)\leq q\}.
\]
Choosing the smallest passing threshold retains the most descriptors under this rule. Every descriptor with $|\lift_{\rm dis}(g)|\geq\tau^*$ moves to holdout, including ties. If no threshold passes, none moves to holdout. We use $q=0.10$, requiring the estimated false discovery proportion at the chosen threshold to be at most $10\%$. This does not require a ten-percentage-point difference in error rates.

The same threshold applies to the whole library. As the threshold changes, both the real and decoy counts change, so their ratio can rise or fall. Testing each descriptor separately at its own lift can, therefore, give different decisions and is not the rule used here.

\paragraph{What the ratio does not guarantee.} The false discovery rate (FDR) is the fraction of false findings averaged across repeated audits. Procedure A does not guarantee that this fraction stays below $q$. With finitely many decoys, $D(\tau)$ can be zero even when differences of that size can arise by chance, so an estimated FDP of zero does not prove that the selected descriptors have genuine error associations. For comparison, we also use exact permutation tests for the same lift statistic. These require errors to be exchangeable across examples under the null. Section~\ref{sec:permutation-baselines} gives the comparison.

\subsection{Checking selected descriptors on holdout}
\label{sec:holdout}

After discovery, we recompute each selected descriptor's lift on the holdout set. We report it only if both groups have sufficient support, the lift has the same sign as in discovery, and $|\lift_{\rm hld}(g)|\geq c$. We choose $c$ before examining holdout errors and use $c=0.10$, requiring an observed difference of at least ten percentage points. A descriptor that passes both discovery and holdout is called \emph{confirmed}.

The thresholds $q$ and $c$ serve different purposes: $q$ controls the estimated discovery ratio, while $c$ sets the minimum observed holdout difference. Confirmation is a reporting rule, not a significance test or a guarantee that the true effect is at least $c$. Failing the rule does not show that no real difference exists and passing it does not establish causality.

\section{Experiments}
\label{sec:experiments}

Our experiments ask two main questions. First, what does \janus{} report on one controlled audit with known structure and on two public benchmarks where the true error patterns are unknown? Second, how does the reporting rule behave under closer statistical checks? We address the second question by comparing standard validation methods, measuring false reports when no associations are present, and testing how often \janus{} detects effects of known size. All three audits use Claude Haiku 4.5 (API identifier \texttt{claude-haiku-4-5-20251001}). The additional analyses reuse the same frozen model outputs and make no new model calls. Appendix~\ref{app:implementation} gives scoring, splits, and filters.\footnote{Code and data: \url{https://github.com/repantis-ai/janus-audit}}

\subsection{Controlled document packets: recovering strong associations}
\label{sec:controlled}

Each controlled example asks for a renewal code in a packet of linked tables. All facts needed to answer are present. A short path maps a project directly to its code, whereas a long path requires project $\to$ vendor $\to$ subsidiary $\to$ renewal code. An indirect question starts from a staff badge and requires two additional lookups to recover the project. Answers are short alphanumeric codes scored by exact match. Appendix~\ref{app:worked} gives a worked example.

The five base descriptors indicate whether the task requires a long lookup chain, uses an indirect question, contains colliding distractor rows, places the needed rows late, or uses flattened formatting. The library also includes five selected combinations. For example, \texttt{long\_x\_indirect} is true when both a long chain and an indirect question are present. This positive control tests whether \janus{} recovers the deliberately introduced long-chain difficulty, not whether it discovers an unknown failure mechanism.

The model answers 102 of 160 examples correctly and \janus{} confirms six descriptors (Table~\ref{tab:controlled}). Five are related to long lookup chains, pointing to a common pattern rather than five separate failure mechanisms. In several errors, the model returns an intermediate vendor or subsidiary key instead of completing the chain (see Appendix~\ref{app:worked}). The remaining descriptor, \texttt{target\_late}, has negative lift: examples with the needed rows placed late have lower error in this audit, which we treat as an association specific to this experiment.

\begin{table}[t]
\centering
\small
\setlength{\tabcolsep}{4pt}
\begin{tabular}{@{}lrrrrr@{}}
\toprule
Descriptor & Error on & Error off & Full lift & Discovery & Holdout \\
\midrule
\texttt{long\_chain} & 65.0\% & 7.5\% & $+0.58$ & $+0.67$ & $+0.48$ \\
\texttt{long\_x\_indirect} & 72.5\% & 24.2\% & $+0.48$ & $+0.55$ & $+0.42$ \\
\texttt{hard\_join\_combo} & 70.0\% & 31.4\% & $+0.39$ & $+0.44$ & $+0.34$ \\
\texttt{long\_x\_collision} & 62.5\% & 27.5\% & $+0.35$ & $+0.27$ & $+0.42$ \\
\texttt{flat\_x\_long} & 57.5\% & 29.2\% & $+0.28$ & $+0.27$ & $+0.30$ \\
\texttt{target\_late} & 22.5\% & 50.0\% & $-0.28$ & $-0.25$ & $-0.30$ \\
\bottomrule
\end{tabular}
\caption{Confirmed controlled-audit descriptors. Error on and Error off use the full audit; Full lift is their difference. Discovery and Holdout are lifts on the separate splits used by the reporting rule. Values are rounded to two decimals. Appendix~\ref{app:worked} defines the combinations.}
\label{tab:controlled}
\end{table}

\subsection{Public benchmarks: no confirmed descriptors}
\label{sec:natural}

MuSiQue \citep{trivedi2022musique} and LongBench v2 \citep{bai2025longbenchv2} test the reporting rule when plausible explanations are available, but whether these are true is unknown. Their metadata provide candidate descriptors, however, neither benchmark tells us which descriptor associations are real. A result with no confirmed descriptors, therefore, means only that none passed our reporting checks, not that the model's errors lack systematic structure.

On MuSiQue, the model answers 84 of 150 questions correctly with all supporting paragraphs in context. Candidates include hop count, the distance between supporting paragraphs, answer position, distractor count, and question length. Some descriptors are too imbalanced for reliable comparison. For example, \texttt{support\_nonadjacent} is true for 141 examples, leaving only nine outside the slice, so it is excluded. None of the seven eligible descriptors passes Procedure A's discovery screen.

For LongBench v2, 115 of the 503 examples in the loaded snapshot meet the recorded answer-format and context-size constraints. All 115 are labeled \texttt{short} in that snapshot, so this is not an evaluation of the benchmark's longer examples. The model answers 56 correctly. Candidates include context size, question type, answer-option overlap, entity and number densities, and domain metadata. None of the 41 eligible descriptors passes discovery. This does not imply that the model's errors are random. As the power analysis below shows, these non-confirmations do not rule out moderate error-rate differences.

\subsection{Comparison with standard validation}
\label{sec:permutation-baselines}

For each descriptor, we ask whether its observed error-rate difference could arise if the descriptor was unrelated to the model's errors. We keep the group sizes fixed and consider every reassignment of the yes/no labels across examples. The exact permutation $p$-value is the fraction of these reassignments that produce a difference at least as large as the observed one, whether the slice has higher or lower error. A small $p$-value means that such a difference is uncommon under this no-association comparison. We first test each descriptor separately at $p\leq0.05$. Because testing many descriptors increases the chance that some look significant even without a real association, we also apply Benjamini--Hochberg (BH) at $q=0.05$ and $q=0.10$ \citep{benjamini1995controlling}. BH ranks all $p$-values and adjusts the selection threshold to account for the number of descriptors tested.

We use 499,999 permutations per audit. In each permutation, we shuffle the errors once and use that same shuffle for every descriptor. The descriptors themselves are left unchanged, preserving which properties tend to occur together. We use these shuffles in two ways. A max-$T$ rule asks whether an observed absolute lift exceeds the largest absolute lift produced in each shuffled audit; we use a 5\% selection level. A mean-count screen estimates how many descriptors exceed a given absolute-lift threshold on average across the shuffles, then chooses one common threshold. Appendix~\ref{app:validation-details} gives the exact calculations.

\begin{table}[t]
\centering
\small
\begin{tabular}{@{}lccc@{}}
\toprule
Discovery rule & Controlled & MuSiQue & LongBench v2 \\
\midrule
Exact permutation, $p\leq0.05$ & 4/4 & 0/0 & 0/0 \\
BH, $q=0.05$ & 2/2 & 0/0 & 0/0 \\
BH, $q=0.10$ & 6/6 & 0/0 & 0/0 \\
Shared max-$T$, 5\% level & 3/3 & 0/0 & 0/0 \\
Shared mean-count screen, $q=0.10$ & 6/6 & 0/0 & 0/0 \\
\janus{} Procedure A, $q=0.10$ & 6/6 & 0/0 & 0/0 \\
\bottomrule
\end{tabular}
\caption{Matched validation comparison. Entries are discovery selections/holdout confirmations. Within each audit, all methods use the same splits, candidate family, lift statistic, and holdout gate. Nominal targets and error criteria differ across rows.}
\label{tab:validation}
\end{table}

Table~\ref{tab:validation} shows that BH at $q=0.10$, the mean-count screen, and Procedure A select the same six controlled descriptors and none on either public benchmark. The other validation rules select fewer controlled descriptors but likewise none on the public benchmarks. The decoy screen shows no advantage over permutation-based validation here, but the controlled findings are not specific to the decoy screen.

\subsection{False reports when no associations are present}
\label{sec:clean-null}

To measure false positives, we keep all descriptor labels fixed and shuffle only the errors within the discovery and holdout sets. This preserves group sizes, descriptor co-occurrence, and the number of errors in each split while breaking descriptor--error associations. Any association found after the shuffle is, by construction, false. We repeat this 50,000 times, regenerating the decoys each time.

\begin{table}[t]
\centering
\small
\begin{tabular}{@{}lrrrr@{}}
\toprule
& \multicolumn{2}{c}{BH, $q=0.10$} & \multicolumn{2}{c}{Procedure A, $q=0.10$} \\
\cmidrule(lr){2-3}\cmidrule(l){4-5}
Audit & Discovery & Confirmed & Discovery & Confirmed \\
\midrule
Controlled & 5.4\% & 1.4\% & 11.0\% & 3.3\% \\
MuSiQue & 6.6\% & 1.6\% & 9.9\% & 2.6\% \\
LongBench v2 & 4.6\% & 1.4\% & 23.0\% & 8.5\% \\
\bottomrule
\end{tabular}
\caption{Probability of at least one selection under the conditional complete null. Every selection is false in this experiment. Same holdout gate for both methods. Each estimate uses 50,000 replays; the Monte Carlo standard error is at most 0.23 percentage points.}
\label{tab:clean-null}
\end{table}

Table~\ref{tab:clean-null} shows that the holdout check reduces false reports for both methods. The number of decoys also matters. On LongBench v2, increasing $K$ from 100 to 1,000 reduces the chance of any discovery from 22.6\% to 9.8\% (uses a different seed from Table~\ref{tab:clean-null}) and the chance of any confirmed finding from 8.4\% to 3.7\%. Appendix~\ref{app:validation-details} gives the full grid. These conditional results do not establish universal error rates or identify a generally safe choice of $K$.

\subsection{Power and detectable effects}
\label{sec:power}

Power is the probability of detecting an effect that is truly present. To estimate it, we run a synthetic simulation using each audit's sample size, discovery/holdout split, numbers of descriptors and decoys, and support rules. We do not simulate the table-lookup, MuSiQue, or LongBench v2 tasks themselves. One simulated descriptor truly increases the error rate by $\delta$: errors occur with probability $0.40$ outside its slice and $0.40+\delta$ inside it. All other descriptors are unrelated to the errors, and each descriptor is present in half the examples. We then apply Procedure A and the same holdout check. Table~\ref{tab:power} reports how often the true descriptor is confirmed.

\begin{table}[t]
\centering
\small
\setlength{\tabcolsep}{4pt}
\begin{tabular}{@{}lcccrrr@{}}
\toprule
Audit configuration & Discovery/holdout & $L$ & $K$ & $\delta=0.05$ & $\delta=0.10$ & $\delta=0.15$ \\
\midrule
Controlled & 80/80 & 10 & 200 & 0.9\% & 3.3\% & 8.7\% \\
MuSiQue & 75/75 & 7 & 100 & 1.0\% & 3.7\% & 9.7\% \\
LongBench v2 & 55/60 & 41 & 100 & 0.4\% & 1.5\% & 3.8\% \\
\end{tabular}
\caption{Confirmation probability for the designated signal in 10,000 repetitions per cell. Effects of $0.05$, $0.10$, and $0.15$ are differences of 5, 10, and 15 percentage points. Confirmation counts either direction when discovery and holdout signs agree.}
\label{tab:power}
\end{table}

With samples this small, \janus{} rarely confirms a single real pattern when the two groups' error rates differ by only 5 to 15 percentage points (Table~\ref{tab:power}). We also simulated larger differences to see when detection becomes more reliable. \janus{} confirms the planted pattern in about 80\% of simulations only when the true difference reaches roughly 36--37 percentage points for the controlled and MuSiQue configurations, and 46--47 points for LongBench. Appendix~\ref{app:validation-details} gives the exact results. The controlled audit contains several related long-chain patterns with much larger effects and, therefore, its successful recovery does not imply that \janus{} can reliably detect a single moderate effect.

The holdout rule adds another limit: a confirmed pattern must show an observed holdout difference of at least 10 percentage points. Smaller true effects are, therefore, not reliably reportable under this rule, even with more data. Larger effects can benefit from larger samples but must still pass discovery. Appendix~\ref{app:validation-details} gives a simple planning approximation. The public-benchmark non-confirmations should be read with these limits in mind.

\subsection{Slice search and the fixed-floor ablation}
\label{sec:ablation}

We use SliceLine \citep{sagadeeva2021sliceline} to show what a slice-search method would surface as candidate error patterns and identify subgroups with high error rates in all three audits: 88--95\% error for controlled data, 55--57\% for MuSiQue, and 85--93\% for LongBench v2. On the public benchmarks, \janus{} confirms none of the tested descriptors, even though SliceLine surfaces high-error groups that could appear worth pursuing. See Appendix~\ref{app:sliceline}.

We also report a fixed-floor ablation (Table~\ref{tab:ablation}). Unlike Procedure A, it uses the 95th percentile of absolute decoy lift as its discovery floor, then applies the holdout gate. This analysis has its own preprocessing, splits, and filters (Appendix~\ref{app:implementation}); it is not a component-by-component ablation of the main Procedure A run.

\begin{table}[t]
\centering
\small
\setlength{\tabcolsep}{4pt}
\begin{tabular}{@{}lrrrrc@{}}
\toprule
Audit & Candidates & Fixed lift & Decoy only & Confirmed & Mean $|\lift|$ \\
\midrule
Controlled & 10 & 8 & 6 & 6 & $0.46\to0.34$ \\
MuSiQue & 7 & 1 & 0 & 0 & n/a \\
LongBench v2 & 34 & 20 & 1 & 0 & $0.36\to0.05$ \\
\bottomrule
\end{tabular}
\caption{Fixed-floor ablation. Within each audit, real-candidate variants share the initial eligibility filter. Fixed lift uses $|\lift_{\rm dis}|\geq0.10$; Decoy only requires a magnitude strictly above the decoy 95th percentile; Confirmed adds holdout. The last column tracks mean absolute lift of decoy survivors from discovery to holdout.}
\label{tab:ablation}
\end{table}

What happens to the original 20 LongBench candidates? We test these candidates on their original ablation split, not the main run's different 41-descriptor setup. Of the 20 with discovery lift magnitude at least $0.10$, one passes an unadjusted exact permutation test at $0.05$; none passes BH at $0.05$ or $0.10$ applied to all 34 eligible candidates. The unadjusted survivor, \texttt{high\_entity\_phrase\_density}, has discovery lift of $-0.3573$, exact $p=0.00825$, BH-adjusted $p=0.28050$, and holdout lift of $-0.0481$. Both select the same candidate, which then fails holdout. The drop in lift from $-0.36$ on discovery to $-0.05$ on holdout is consistent with selection exaggerating the initial effect, but it does not show that the true association is zero. Applying only the holdout gate to the 20 fixed-lift candidates leaves three, whose validity is unknown. The ablation supports explicit validation and replication, rather than a unique advantage of decoys.

\section{Related work}
\label{sec:related}

\paragraph{Slice discovery and behavioral testing.} SliceFinder searches high-error feature conjunctions and uses alpha-investing during search \citep{chung2019slicefinder}; SliceLine uses linear algebra for scalable enumeration \citep{sagadeeva2021sliceline}. Domino, Spotlight, InfEmbed, and MCSD use learned representations or geometry to find coherent error regions \citep{eyuboglu2022domino,deon2022spotlight,wang2023infembed,yu2026mcsd}. LADDER and HiBug2 use language or generated attributes to propose hypotheses \citep{ghosh2025ladder,chen2025hibug2}. CheckList and AdaTest organize or expand behavioral tests \citep{ribeiro2020checklist,ribeiro2022adatest}. These are potential sources of candidates, provided outcome-informed proposal and subsequent validation are separated. Long-context studies and benchmarks likewise reveal useful task structure without resolving which selected explanations merit reporting \citep{liu2024lost,bai2025longbenchv2}.

\paragraph{Subgroup auditing.} Fairness auditing also asks whether errors or disparities concentrate on subpopulations. Cherian and Cand\`es use bootstrap calibration for simultaneous guarantees in model-agnostic fairness auditing \citep{cherian2024fairness}. AFISP identifies poorly performing clinical subsets and interpretable phenotypes \citep{subbaswamy2024afisp}. FairTree studies subgroup performance with permutation and fluctuation-test variants \citep{debelak2026fairtree}. \janus{} shares this motivation but studies a specific reporting rule for named error descriptors.

\paragraph{Manufactured nulls and multiple testing.} Knockoffs and target-decoy methods compare observations with constructed controls and obtain guarantees under their respective assumptions \citep{barber2015knockoffs,candes2018knockoffs,elias2007targetdecoy}. Related chance comparisons appear in retrieval evaluation \citep{repantis2026bor}. BH provides a standard multiple-testing procedure for valid $p$-values under suitable dependence conditions \citep{benjamini1995controlling}. \janus{} borrows the idea of manufactured comparisons, not those formal guarantees. For the fixed Boolean descriptors studied here, exact permutation $p$-values are available under the assumptions of the permutation test described above. Our comparison, therefore, asks how the finite decoy screen behaves relative to standard validation, not whether standard testing is possible.

\section{Limitations}
\label{sec:limitations}

\janus{} evaluates a proposed library, not the full space of explanations. It cannot recover a failure mode that was not proposed and confirmation establishes association rather than causation. Decoy shuffles do not preserve descriptor co-occurrence. Our experiments use one language model and modest samples, including only the feasible short-length subset of LongBench v2. The power simulation uses balanced, independent descriptors, one signal, and a single baseline error rate, so it does not characterize rare slices, correlated signals, adaptive proposal systems, or all audit sizes. We do not establish superiority over BH.

Procedure A has no finite-sample FDR guarantee and the clean-null results depend substantially on the decoy count. The replays hold descriptor matrices and error totals fixed; they do not establish control with non-null hypotheses or clustered observations. The holdout gate is not a significance test. Filtering an FDR-controlled discovery set does not automatically preserve FDR for the final subset, so any formal guarantee must cover the complete reporting procedure. Holdout errors must not be reused to tune candidates or thresholds. Repeated splits of frozen outcomes assess sensitivity, not independent confirmation.

\section{Conclusion}
\label{sec:conclusion}

A plausible error pattern is not automatically a reportable finding. \janus{} turns that distinction into an explicit audit procedure: fix the candidate properties, compare their discovery effects with decoys, and check survivors on holdout. It recovers the strong controlled associations and reports none on the two public benchmarks, matching permutation testing with BH. This agreement connects the decoy screen to established validation, while the null and power experiments reveal where the procedure is less reliable: finite decoy counts affect false reporting and these audit sizes rarely confirm moderate isolated effects. The broader contribution is a transparent way to separate candidate explanations from findings, non-confirmation from absence, and empirical screening from formal error control.

\bibliographystyle{colm2026_conference}
\bibliography{janus}

\clearpage
\appendix

\section{A worked example of the controlled task}
\label{app:worked}

This example is based on a recorded long-chain error from the controlled audit. The answer path and model response are preserved: the model returned \texttt{KAPPA-759} instead of \texttt{QZ-4421}. Surrounding rows are illustrative distractors added for readability. The actual audit packet contains many more rows.

\paragraph{The packet.} Each example contains plain-text tables linked by shared keys. The model is instructed to use only these tables and return one renewal code. The trimmed packet below shows the answer path and a few distractors.

\begin{quote}\small\ttfamily
STAFF DIRECTORY\\
badge | name | division\\
BD-40271 | Dana Ives | Cobalt\\
BD-55810 | Pat Orr | Aurora\\
\ldots\\[2pt]
ASSIGNMENT LEDGER\\
name | division | project\\
Dana Ives | Cobalt | HARBOR-118\\
Pat Orr | Aurora | SUMMIT-204\\
\ldots\\[2pt]
PROJECT TO VENDOR CROSSWALK\\
project | vendor\\
HARBOR-118 | KAPPA-759\\
SUMMIT-204 | VANTA-031\\
\ldots\\[2pt]
VENDOR TO SUBSIDIARY CROSSWALK\\
vendor | subsidiary\\
KAPPA-759 | SUBD-130\\
VANTA-031 | SUBA-907\\
\ldots\\[2pt]
SUBSIDIARY RENEWAL REGISTRY\\
subsidiary | renewal\_code\\
SUBD-130 | QZ-4421\\
SUBA-907 | RX-8890\\
\ldots
\end{quote}

\paragraph{The question and answer path.} The question is: ``What is the renewal code for project \texttt{HARBOR-118}?'' First find the project's vendor, then the vendor's subsidiary, then the subsidiary's renewal code:
\[
\texttt{HARBOR-118}\;\to\;\texttt{KAPPA-759}\;\to\;\texttt{SUBD-130}\;\to\;\texttt{QZ-4421}.
\]
Each arrow is one table lookup. The correct answer is the final code, \texttt{QZ-4421}, not either intermediate key.

\paragraph{What the model returned.} The answer was \texttt{KAPPA-759}, the vendor found after the first lookup. The response is consistent with stopping early along the chain. It does not reveal the model's internal reasoning, but it shows that the returned key is an intermediate result rather than the requested renewal code.

\begin{figure}[t]
\centering
\begin{tikzpicture}[key/.style={draw,rounded corners,align=center,inner sep=4pt,font=\small\ttfamily,minimum height=0.75cm},role/.style={font=\small,text=gray!55!black},arr/.style={-Latex,thick},lbl/.style={font=\small\itshape,text=gray!45!black,midway,above=4pt}]
\node[key,fill=gray!8] (p) {HARBOR-118};
\node[key,fill=blue!7,right=1.25cm of p] (v) {KAPPA-759};
\node[key,fill=blue!7,right=1.25cm of v] (s) {SUBD-130};
\node[key,fill=green!12,right=1.25cm of s] (c) {QZ-4421};
\node[role,below=2pt of p] {project};
\node[role,below=2pt of v] {vendor};
\node[role,below=2pt of s] {subsidiary};
\node[role,below=2pt of c] {renewal code};
\draw[arr] (p) -- node[lbl] {lookup 1} (v);
\draw[arr] (v) -- node[lbl] {lookup 2} (s);
\draw[arr] (s) -- node[lbl] {lookup 3} (c);
\node[font=\small,text=red!65!black,align=center,below=1.1cm of v] (stop) {returned key:\\ \texttt{KAPPA-759}};
\draw[-Latex,red!55!black,dashed] (stop) -- (v);
\end{tikzpicture}
\caption{The three lookups in the worked example. Each box is a key. The returned vendor key is the first intermediate result; the requested answer is the final renewal code.}
\label{fig:chain-app}
\end{figure}

\paragraph{How the five factors vary the task.} With \texttt{long\_chain} off, one registry maps project directly to renewal code; with it on, the path with three lookups above replaces that registry. With \texttt{indirect\_query} on, the question names a staff badge instead of a project, adding badge $\to$ name/division $\to$ project before the registry lookup. \texttt{collision\_distractors} introduces rows sharing a staff name or a similar name, so a name match alone can be ambiguous. \texttt{target\_late} moves needed rows near the ends of their tables. \texttt{flat\_format} renders tables as running \texttt{field=value} text rather than separated rows.

\paragraph{Combination descriptors.} The five additional descriptors are conjunctions of the base factors. \texttt{long\_x\_indirect} requires a long chain and an indirect query; \texttt{collision\_x\_indirect} requires collisions and an indirect query; \texttt{long\_x\_collision} requires a long chain and collisions; \texttt{hard\_join\_combo} requires all three; and \texttt{flat\_x\_long} requires a long chain in flattened format. A confirmed conjunction is an association for that combined slice, not evidence of an interaction effect beyond its constituent factors.

\FloatBarrier
\section{Detailed SliceLine outputs}
\label{app:sliceline}

Table~\ref{tab:sliceline_examples} gives representative slice-search results. They use the same evaluated outcomes as the audit, but not the same filtered candidate family or discovery protocol as the matched validation comparison. For example, SliceLine can use \texttt{support\_nonadjacent} even though the main MuSiQue reporting rule filters it out. These outputs illustrate candidate subgroups; they are not confirmed findings under \janus{}.

\begin{table}[htbp]
\centering
\small
\begin{tabularx}{\linewidth}{@{}l>{\raggedright\arraybackslash}Xr@{}}
\toprule
Audit & Slice & Error \\
\midrule
Controlled & \texttt{long\_chain=1} AND \texttt{target\_late=0} & 88\% \\
Controlled & \texttt{flat\_format=0} AND \texttt{long\_chain=1} AND \texttt{target\_late=0} & 95\% \\
MuSiQue & \texttt{support\_nonadjacent=1} AND \texttt{high\_question\_words=0} & 55\% \\
LongBench v2 & \texttt{high\_context\_words=0} AND \texttt{high\_question\_context\_overlap=1} AND \texttt{high\_entity\_phrase\_density=0} & 93\% \\
\bottomrule
\end{tabularx}
\caption{Representative SliceLine outputs. A value of 0 means that the corresponding property is absent. Long conjunctions are search results, not separate evidence that each constituent property explains an error.}
\label{tab:sliceline_examples}
\end{table}

\FloatBarrier
\section{Validation and sensitivity details}
\label{app:validation-details}

\subsection{Exact permutation tests}

Let $n_d$ be discovery size, $E$ its error total, $m$ the number of examples with a descriptor, and $A$ the observed errors among those examples. Holding $E$ and $m$ fixed while assigning the descriptor to examples uniformly at random gives
\[
A'\sim\operatorname{Hypergeometric}(n_d,E,m),\qquad \Delta(a)=\frac{a}{m}-\frac{E-a}{n_d-m}.
\]
The hypergeometric distribution counts how many of the $E$ errors fall in a random group of size $m$. We sum the probabilities of all possible counts whose lift magnitude is at least the observed magnitude:
\[
p_g=\Pr\{|\Delta(A')|\geq|\Delta(A)|\mid n_d,E,m\}.
\]
This is the inclusive absolute-lift tail, not a mid-$p$ value or a doubled one-sided convention. Its validity requires the stated exchangeability null. BH is applied to all eligible descriptors, not only those with large observed lifts. For ordered $p$-values $p_{(1)}\leq\cdots\leq p_{(L)}$, it finds the largest $r$ with $p_{(r)}\leq qr/L$ and selects the first $r$ hypotheses; it selects none when there is no such $r$.

\subsection{Shared permutations and pooled decoys}

For each shared permutation $b$, let $T_{bj}$ be the null lift for descriptor $j$, and let $T_j$ be its observed discovery lift. The max-$T$ comparison uses $M_b=\max_j|T_{bj}|$ and the adjusted Monte Carlo value
\[
p_j^{\max}=\frac{1+\sum_{b=1}^{B}\ind{M_b\geq|T_j|}}{B+1},\qquad B=499{,}999.
\]
We select candidates at $p_j^{\max}\leq0.05$. This is a familywise comparison under the complete permutation null. Strong control when only some descriptor nulls are true requires additional conditions, such as the relevant null distributions being unchanged by which other effects are present. We do not claim those conditions for arbitrary audit libraries.

The shared mean-count screen instead uses
\[
\widehat{V}_{\rm shared}(\tau)=\frac{1}{B}\sum_{b=1}^{B}\sum_{j=1}^{L}\ind{|T_{bj}|\geq\tau},\qquad \widehat{\mathrm{FDP}}_{\rm shared}(\tau)=\frac{\widehat{V}_{\rm shared}(\tau)}{\max(1,R(\tau))}.
\]
It chooses one most-permissive passing threshold, exactly as Procedure A does. It remains an empirical screen rather than a further FDR guarantee.

Preserving descriptor dependence changes the joint null distribution, which matters for max-$T$. It does not change the sum of marginal tail probabilities at a fixed threshold:
\[
\mathbb{E}\!\left[\sum_{j=1}^{L}\ind{|T_j^\pi|\geq\tau}\right]=\sum_{j=1}^{L}\Pr(|T_j^\pi|\geq\tau),
\]
where $\pi$ denotes a shared row permutation. When independent decoy sources are sampled uniformly from that same eligible family, the expected scaled count $(L/K)D(\tau)$ is the same marginal-tail sum. Thus the shared mean-count and pooled-decoy screens do not target different population mean counts merely because one preserves correlations. Their finite simulation variability can differ. This identity at a fixed threshold does not justify FDR control after choosing a threshold from the data.

\subsection{What complete-null error measures}

Under the conditional complete-null law, every finding is false. If $R_{\rm out}$ is the number of findings, its false-discovery proportion is $\ind{R_{\rm out}>0}$. The probability of any finding therefore equals both the familywise error rate and FDR \emph{in this complete-null experiment}. The identity does not extend to a mixture of true and false associations. Table~\ref{tab:clean-null} measures separate probabilities for a nonempty discovery set and a nonempty confirmed set. It does not estimate the truth status of descriptors on the original public benchmarks.

\subsection{Sensitivity to the number of decoys}

For the LongBench v2 descriptor matrix, we run 50,000 paired complete-null replays with seed 90260812. Within each replay, discovery and holdout errors are shuffled independently and 1,000 decoys are generated in order. Smaller choices of $K$ use prefixes of those same decoys. All configurations therefore share the same shuffled outcomes and nested decoy draws.

\begin{table}[htbp]
\centering
\small
\begin{tabular}{@{}rrrr@{}}
\toprule
$K$ & $K/L$ ($L=41$) & Any discovery & Any confirmation \\
\midrule
41 & 1.00 & 40.4\% & 16.8\% \\
100 & 2.44 & 22.6\% & 8.4\% \\
200 & 4.88 & 14.1\% & 5.2\% \\
500 & 12.20 & 12.5\% & 4.6\% \\
1,000 & 24.39 & 9.8\% & 3.7\% \\
\bottomrule
\end{tabular}
\caption{LongBench v2 complete-null sensitivity to decoy count. All rows use the common-threshold Procedure A. These are conditional replay probabilities, not guaranteed error bounds for these values of $K$.}
\label{tab:decoy-count}
\end{table}

The $K=100$ result differs slightly from Table~\ref{tab:clean-null} because the paired experiment uses a different seed and random-number construction. Increasing $K$ reduces false reporting on this grid, but the experiment does not prove monotonicity, convergence to BH, or a universally adequate ratio $K/L$. It also does not change the $K=100$ configuration used in the observed LongBench audit.

\subsection{Power simulation and effect probes}

Each simulated descriptor membership is independently Bernoulli$(0.5)$. One designated descriptor changes error probability from $p_0=0.4$ to $p_0+\delta$; errors are drawn independently given membership. We apply the coded count minima to both splits before running Procedure A and the holdout rule. This preserves the main configurations' sample sizes, candidate counts, and decoy counts, not their actual metadata construction or full prevalence structure. Ineligible instances of the signal descriptor count as non-confirmations. The metric counts confirmation whenever both observed signs agree and the holdout magnitude reaches $0.10$, including rare agreements in the direction opposite to the positive population effect.

\begin{table}[htbp]
\centering
\small
\begin{tabular}{@{}lrrrr@{}}
\toprule
Configuration & Lower effect & Confirmation & Upper effect & Confirmation \\
\midrule
Controlled & 0.36 & 78.84\% & 0.37 & 81.46\% \\
MuSiQue & 0.36 & 77.17\% & 0.37 & 80.24\% \\
LongBench v2 & 0.46 & 79.67\% & 0.47 & 82.13\% \\
\bottomrule
\end{tabular}
\caption{Additional power probes, each based on 10,000 repetitions. These are five tested effects per configuration when combined with Table~\ref{tab:power}, not a dense power curve. The point estimates straddle 80\%; the endpoints are not confidence bounds on a minimum detectable effect.}
\label{tab:power-probes}
\end{table}

For the controlled and MuSiQue configurations, tested effects are $0.05,0.10,0.15,0.36,0.37$; for LongBench they are $0.05,0.10,0.15,0.46,0.47$. Seeds are $50260812+100000d+\lfloor10000\delta\rfloor$, where $d=0,1,2$ indexes controlled, MuSiQue, and LongBench. Each cell uses 10,000 independent repetitions. Near an 80\% probability, the Monte Carlo standard error is about 0.4 percentage points. In particular, an estimate of 80.24\% is not a precise demonstration that population power exceeds 80\%. The separate 50,000-replay complete-null experiments use seeds $80260812+100000d$; shared-permutation baseline seeds are $20260812+d$.

\subsection{Approximate holdout planning}

A simple approximation explains why effect size and support matter even before considering discovery. Let $n_h$ be holdout size, $\pi$ the fraction of holdout examples in the slice, $p_0$ the population error probability outside it, and $\delta>0$ the population lift. For independent examples and fixed group sizes, the approximate variance of the holdout difference is
\[
s_h^2=\frac{(p_0+\delta)(1-p_0-\delta)}{n_h\pi}+\frac{p_0(1-p_0)}{n_h(1-\pi)}.
\]
The two terms are the sampling variances of the on-slice and off-slice error rates. Small support on either side makes the difference harder to estimate.

Let $\Phi$ be the standard normal cumulative distribution function. For a positive-direction gate with threshold $c$, the normal approximation is
\[
\Pr(\widehat{\delta}_h\geq c)\approx1-\Phi\!\left(\frac{c-\delta}{s_h}\right).
\]
This alone is not an upper bound for the implemented either-direction confirmation rule. The exact event inclusion is
\[
\Pr(\text{confirmation})\leq\Pr(|\widehat{\delta}_h|\geq c),
\]
whose normal approximation is
\[
\Pr(|\widehat{\delta}_h|\geq c)\approx1-\Phi\!\left(\frac{c-\delta}{s_h}\right)+\Phi\!\left(\frac{-c-\delta}{s_h}\right).
\]
The inequality concerns exact probabilities; the expression involving $\Phi$ is only an approximation, especially for small groups. Neither calculation includes the discovery screen.

For a desired positive-direction holdout pass probability $1-\beta>1/2$, with $\delta>c$ and normal quantile $z_{1-\beta}$, rearranging gives
\[
n_h\gtrsim\frac{z_{1-\beta}^2}{(\delta-c)^2}\left[\frac{(p_0+\delta)(1-p_0-\delta)}{\pi}+\frac{p_0(1-p_0)}{1-\pi}\right].
\]
This is a planning approximation for the holdout gate, not a sample-size guarantee for the full audit. At $\delta=c$, the regular normal approximation gives a positive-direction pass probability near one half, not 80\%. For any fixed $|\delta|<c$, consistency makes the absolute-gate pass probability tend to zero. Choosing the observed-effect floor is therefore part of specifying what effects the audit is intended to report.

\FloatBarrier
\section{Implementation notes}
\label{app:implementation}

\subsection{Frozen evaluations and scoring}

All three audits evaluate Claude Haiku 4.5 (\texttt{claude-haiku-4-5-20251001}) with constrained short-answer or multiple-choice outputs. The controlled audit has 160 outcomes and 58 errors. Its factor memberships and correctness can be reconstructed from the saved execution log; that reconstruction is not a replacement for original prompts or raw responses. Controlled correctness uses exact match to the requested renewal code.

MuSiQue has 150 outcomes and 66 errors. Its recorded scorer lowercases the prediction and gold answers, removes articles, replaces characters other than letters, digits, and whitespace, and collapses whitespace. It accepts a normalized match to the gold answer or an alias, a prediction containing a normalized gold answer with at most three extra tokens, or token F1 of at least $0.8$ against a normalized gold answer or alias. The audit error indicators use that scorer, not a new exact-match-only reevaluation. All supporting paragraphs are provided in context.

The loaded LongBench v2 snapshot contains 503 examples. Eligibility requires a gold answer in A, B, C, or D, four nonempty answer options, no more than 30,000 context words counted by the notebook's regular expression, and no more than 120,000 context characters. The saved run evaluates all 115 eligible examples without oversampling. All are marked \texttt{short} in that snapshot. The model returns a letter through the constrained answer tool; correctness is equality to the gold letter. There are 59 errors. These fixed outputs supply the camera-ready comparisons.

\subsection{Main splits, descriptors, and support}

\begin{table}[htbp]
\centering
\small
\begin{tabular}{@{}lrrrrr@{}}
\toprule
Audit & Total $n$ & Discovery & Holdout & Eligible $L$ & Decoys $K$ \\
\midrule
Controlled & 160 & 80 & 80 & 10 & 200 \\
MuSiQue & 150 & 75 & 75 & 7 & 100 \\
LongBench v2 & 115 & 55 & 60 & 41 & 100 \\
\bottomrule
\end{tabular}
\caption{Main configurations used for Procedure A and the matched validation comparison. Historical SliceLine and fixed-floor ablation settings differ as described below.}
\label{tab:configuration}
\end{table}

Controlled outcomes are shuffled with Python's \texttt{random.Random(11)} and split 80/80. The ten descriptors are the five factors and combinations in Appendix~\ref{app:worked}. Each eligible descriptor has at least eight examples on each side in each split, corresponding to prevalence between $0.10$ and $0.90$ for these split sizes.

MuSiQue is shuffled with \texttt{random.Random(7)} and split 75/75. Binary metadata are retained; numeric metadata are converted to indicators of being strictly above the discovery median. Thresholds are applied unchanged to holdout. Seven descriptors pass full-audit prevalence in $[0.10,0.90]$, at least three on-slice examples per split, and nonempty complements: \texttt{high\_num\_support}, \texttt{high\_num\_hops}, \texttt{high\_support\_distance}, \texttt{answer\_para\_late}, \texttt{high\_distractor\_count}, \texttt{high\_answer\_alias\_count}, and \texttt{high\_question\_words}. In the saved metadata, \texttt{support\_nonadjacent} means a gap of at least two paragraph indices between the earliest and latest supporting paragraphs; \texttt{answer\_para\_late} means that the latest supporting paragraph lies at or beyond 70\% of the paragraph list. These are metadata proxies, not observations of the model's reasoning.

LongBench v2 is stratified by subdomain in recorded insertion order. Using \texttt{random.Random(42)}, each stratum is shuffled and the floor of half its examples goes to discovery, giving 55/60 overall. Numeric thresholds are discovery medians and categorical descriptors use levels seen in discovery. Forty-one descriptors satisfy at least three on-slice examples in each split, at least eight on-slice examples overall, and nonempty complements. These public-benchmark support minima are not the stricter eight-on/eight-off filter of the fixed-floor ablation.

\subsection{Decoys and the common-threshold implementation}

Main runs use $q=0.10$ and the decoy counts in Table~\ref{tab:configuration}. Controlled decoys continue the Python random-number-generator state used for splitting. Public-benchmark decoys restart their respective seeded generators after the split. The LongBench decoy pool additionally checks discovery prevalence in $[0.05,0.95]$; all 41 eligible main-run descriptors satisfy it. The historical implementation also shuffles unused holdout decoy memberships. Those shuffles do not contribute statistics but consume random numbers, so exact replay retains them.

The final Procedure A implementation uses the single common threshold defined in Section~\ref{sec:procedure-a}. Earlier notebook cells marked a descriptor by evaluating the estimated ratio at its own lift. Because the ratio is not monotone, these rules can disagree. Replacing the earlier decisions with the common-threshold rule leaves the three observed main finding sets unchanged, but changes some simulation and repeated-split results. All Procedure A simulation tables here use the common-threshold results. The per-descriptor ratio saved in a result table is not an adjusted $p$-value and need not itself pass $q$ for every descriptor above the chosen common threshold.

\subsection{Historical fixed-floor ablation and SliceLine}

The \janus{}-lite ablation first binarizes metadata using full-audit medians and category levels, then splits with NumPy \texttt{RandomState(11)}. Initial real-candidate eligibility requires at least eight on-slice and eight off-slice discovery examples and discovery prevalence in $[0.10,0.90]$. Holdout support and prevalence are checked again before confirmation. Decoys use \texttt{RandomState(12)} and a source pool filtered by discovery prevalence; that source pool need not equal the support-eligible real family. These settings are preserved to explain the submitted ablation, not presented as identical to the main analysis.

For LongBench, this produces a 57/58 split, 34 eligible real candidates, a 40-descriptor decoy source pool, and a 95th-percentile absolute-decoy-lift floor of $0.322252$ with $K=200$. A descriptor must be strictly above that floor. These are the settings behind the 20-to-1-to-0 sequence. The comparison of the original 20 candidates uses this same split and tests BH over all 34 eligible descriptors. The historical SliceLine comparison likewise uses full-audit binarization and searches a different candidate space, including descriptors excluded by the main reporting filter.

\subsection{Repeated-split sensitivity}

We repeat the offline analysis for 200 split seeds $20260524+1009s$, with $s=0,\ldots,199$, on the same frozen outputs. The sensitivity implementation sorts LongBench subdomain names before stratification, unlike the main run, and uses a separate decoy-seed offset of 10,000,000. Controlled data require eight examples on each side in each split; the public-benchmark checks require at least three on-slice examples per split, with their other filters recorded in the released analysis. This is a distinct split-sensitivity protocol, not 200 new model evaluations.

Under the common-threshold rule, the controlled audit has a nonempty confirmed set in all 200 splits. MuSiQue is empty in 197/200 and LongBench v2 in 194/200. The six main controlled descriptors recur as follows: \texttt{long\_chain}, 200; \texttt{long\_x\_indirect}, 198; \texttt{long\_x\_collision}, 179; \texttt{target\_late}, 144; \texttt{flat\_x\_long}, 138; and \texttt{hard\_join\_combo}, 132. \texttt{indirect\_query} additionally confirms in two splits. These recurrence counts include ineligible splits in the denominator. Nonempty public-benchmark splits involve descriptors recurring in at most six of the 200 splits. The counts are not complete-null error estimates or evidence of fresh replication.

\paragraph{Cost of decoys.} Generating and scoring $K$ decoys costs $O(Kn)$ once descriptor memberships and errors are available. Each decoy is a shuffled Boolean vector and its score uses two empirical error rates. This cost statement excludes candidate construction, threshold-search implementation, the shared-permutation baseline, and model evaluation. We do not report a timing comparison with exact permutation testing.

\end{document}